\documentclass[sigconf]{acmart}

\usepackage[utf8]{inputenc} 
\usepackage[T1]{fontenc}    
\usepackage{hyperref}       
\usepackage{url}            
\usepackage{booktabs}       
\usepackage{amsfonts}       
\usepackage{nicefrac}       
\usepackage{microtype}      
\usepackage{amsmath}
\usepackage{array}
\usepackage{graphicx}
\usepackage{subcaption}
\usepackage{xspace}
\usepackage{enumitem}
\usepackage{multirow}
\usepackage{subcaption}

\newcommand{\Poincare}{Poincar\'e\xspace}
\newcolumntype{L}[1]{>{\raggedright\let\newline\\\arraybackslash\hspace{0pt}}m{#1}}
\newcommand{\zbf}{\mathbf{z}}
\newcommand{\Zbf}{\mathbf{Z}}
\newcommand{\ebf}{\mathbf{e}}
\newcommand{\Qbf}{\mathbf{Q}}
\newcommand{\Kbf}{\mathbf{K}}
\newcommand{\Vbf}{\mathbf{V}}
\newcommand{\Wbf}{\mathbf{W}}
\newcommand{\Cbf}{\mathbf{C}}
\newcommand{\Xbf}{\mathbf{X}}
\newcommand{\Dbf}{\mathbf{D}}
\newcommand{\Gbf}{\mathbf{G}}

\newcommand{\Hbf}{\mathbf{H}}
\newcommand{\Abf}{\mathbf{A}}

\newtheorem*{theorem*}{Theorem}

\copyrightyear{2020}
\acmYear{2020}
\setcopyright{acmcopyright}
\acmConference[WSDM '20]{The Thirteenth ACM International Conference on Web Search and Data Mining}{February 3--7, 2020}{Houston, TX, USA}
\acmBooktitle{The Thirteenth ACM International Conference on Web Search and Data Mining (WSDM '20), February 3--7, 2020, Houston, TX, USA}
\acmPrice{15.00}
\acmDOI{10.1145/3336191.3371778}
\acmISBN{978-1-4503-6822-3/20/02}

\begin{document}

\title{Product Knowledge Graph Embedding for E-commerce}

\author{Da Xu}
\authornote{Both authors contributed equally to this research.}
\author{Chuanwei Ruan}
\authornotemark[1]
\affiliation{%
  \institution{Walmart Labs}
  \city{Sunnyvale}
  \state{California}
  \country{USA}
}
\email{{Da.Xu,Chuanwei.Ruan}@walmartlabs.com}

\author{Evren Korpeoglu, Sushant Kumar,}
\author{Kannan Achan}
\affiliation{%
  \institution{Walmart Labs}
  \city{Sunnyvale}
  \state{California}
  \country{USA}
}
\email{{EKorpeoglu,SKumar4,KAchan}@walmartlabs.com}


\begin{abstract}
In this paper, we propose a new product knowledge graph (PKG) embedding approach for learning the intrinsic product relations as product knowledge for e-commerce. We define the key entities and summarize the pivotal product relations that are critical for general e-commerce applications including marketing, advertisement, search ranking and recommendation. We first provide a comprehensive comparison between PKG and ordinary knowledge graph (KG) and then illustrate why KG embedding methods are not suitable for PKG learning.
We construct a self-attention-enhanced distributed representation learning model for learning PKG embeddings from raw customer activity data in an end-to-end fashion. We design an effective multi-task learning schema to fully leverage the multi-modal e-commerce data. The \Poincare embedding is also employed to handle complex entity structures. We use a real-world dataset from \textsl{grocery.walmart.com} to evaluate the performances on knowledge completion, search ranking and recommendation. The proposed approach compares favourably to baselines in knowledge completion and downstream tasks.

\end{abstract}



\maketitle

\section{Introduction}
\label{sec:introduction}
\begin{figure}[hbt]
    \centering
    \includegraphics[scale=0.27]{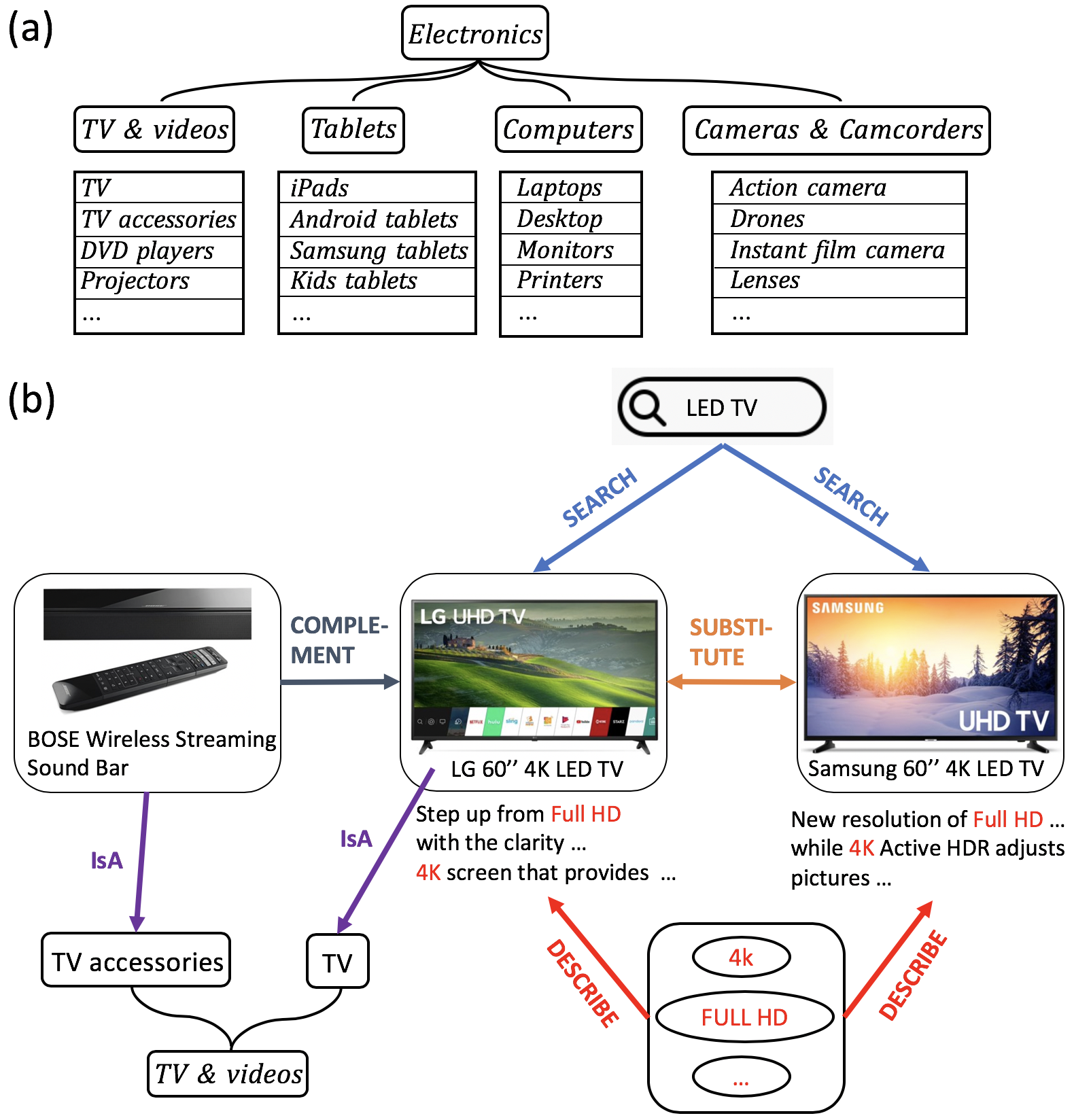}
    \caption{Visual illustrations. (a) Example of product category hierarchy. (b) Sketched product knowledge graph.}
    \label{fig:PKG}
\end{figure}

Understanding the relations among products as product knowledge play pivotal roles in the rapidly developing e-commerce world. Product relations, including \texttt{complement} (\texttt{co-buy}), \texttt{co-view} and \texttt{substitute}, are central for marketing, advertising and recommendation. Several papers have devoted to learning and inferring product relations from either predefined product relation graph \cite{mcauley2015inferring} or customer records \cite{zhang2018quality}. Besides the above relations, the interactions between products and natural language is gaining increasing attention \cite{anelli2018knowledge}. Firstly, product descriptions provide valuable side information for various tasks. Secondly, customers often engage with products via search activities. We use the \texttt{search} and \texttt{describe} relation to summarizing interactions between natural language and products. On top of the descriptions, products are often grouped into hierarchical categories shown in Figure \ref{fig:PKG}a, which motivates the \texttt{IsA} relationship. 
In this paper, we focus on the above six key relations for product knowledge, which should satisfy most e-commerce applications. Notice that these relations can be subdivided into more fine-grained levels depending on the use case. For example, \texttt{complement} can be divided into \texttt{AddOn}, \texttt{AccessoryTo}, \texttt{PartOf}, and \texttt{describe} summarizes \texttt{HasAttribute}, \texttt{Brand}, \texttt{Name}, etc. 
 
By treating products, words and category labels as entities and relations as edges, the multi-relation product knowledge can be efficiently summarized by product knowledge graph like Figure \ref{fig:PKG}b. This motivates us to compare our work for learning PKG with the well-established work in knowledge graph (KG) learning, especially knowledge graph embedding \cite{wang2017knowledge}. KG embedding methods learn the representation (embedding) of entities and relations in a lower-dimensional continuous vector space. The inherent structure of the KG is geometrically preserved in the vector space, and the embeddings can simplify manipulations while remaining useful for downstream applications such as knowledge completion. The prevalent KG embedding models including \textsl{TransE} \cite{bordes2013translating}, \textsl{TransH} \cite{wang2014knowledge}, \textsl{TransR} \cite{lin2015learning} and \textsl{TransD} \cite{ji2015knowledge} measure the plausibility of observed facts in KG with translational distance, while \textsl{RESCAL} \cite{nickel2011three}, \textsl{DistMult} \cite{yang2014embedding}, \textsl{HolE} \cite{nickel2016holographic} and \textsl{ComplEx} \cite{trouillon2016complex} use latent semantic similarity. Subsequent work improves the model complexities to further explore additional information and structural properties of KG \cite{wang2017knowledge}. 


However, most KG embedding models rely on the crucial assumption that all the facts in knowledge base are established with high plausibility, which is hardly valid when learning PKG embedding. In e-commerce data, other than the obvious \texttt{describe} and \texttt{IsA} relation, the \texttt{complement}, \texttt{co-view}, \texttt{substitute} and \texttt{search} relations all need to be extracted from customer records by information retrieval methods.
For example, 
two \texttt{coffee} products that are very similar in the title and brand may not \texttt{substitute} each other because one is \textsl{decaffeinated} and the other is \textsl{caffeinated}. The customer substitution records, in this case, is a better source for informing the \texttt{substitute} relation.
Just like many other machine learning applications in e-commerce, learning PKG embedding cannot circumvent the \textbf{noise} and \textbf{sparsity} issues in the customer-product interaction data. If retrieving target relations is not complicated enough, the irregular tree-structured product categories pose further challenges since they are inherently hard to be embedded into Euclidean spaces.
Last but not least, even the product descriptions are sometimes filled with \textbf{noise} words that are irrelevant to the \texttt{describe} relation. 

Firstly, to break down the \textbf{noise} issues, we point out that our learning objectives are intrinsically discrete event-sequence learning similar to that of the neural machine language translation \cite{bahdanau2014neural}. In the following example, we try to retrieve the \texttt{complement} relation for \textsl{toothpaste} and \textsl{toothbrush} from a purchase sequence:
\begin{equation*}
    [\ldots,\text{soap}, \text{detergent}, \underset{\uparrow}{\text{toothbrush}},{\text{towel}} \to \text{toothpaste}, \ldots].
\end{equation*}
Among the products purchased prior to \textsl{toothpaste}, the \textsl{toothbrush} should be recognized with highest importance such that the input product sequence 'translates' to \textsl{toothpaste}. Same argument applies to product description, where for an ice cream described by:
\begin{equation*}
\begin{aligned}
    \{\text{The } \underset{\uparrow}{\text{strawberry }} \text{ice cream } \text{featured by } \underset{\uparrow}{\text{Haagan-Dazs }} \text{is } \text{the } \\
    \text{marriage } \text{of } \text{sweet } \text{summer } \text{strawberries } \text{to } \text{cream } \text{and } \ldots \},
\end{aligned}
\end{equation*}
the flavor and brand should receive the highest attentions such that the description 'translates' to \textsl{strawberry Haggan-Dazs ice cream}. It becomes clear at this point that our goal is aligned with the attention mechanism proposed in neural machine translation literature \cite{luong2015effective,vaswani2017attention}, where a sentence is represented by a weighted sum of the individual word representation. The attention weight assigned to each entry reflects their relative importance when translating the next token. Therefore, we propose an adapted self-attention network as the relation extractor to effectively retrieve target relations from the noisy product description and customer activity data. 

Secondly, we resort to the multi-task learning with multi-modal data to handle the \textbf{sparsity} issue. It has become a common practice for industrial applications to leverage the useful information across related tasks to make up for the data sparsity in individual task \cite{wang2018multi,zhou2018deep,ma2018modeling}. In e-commerce, available data sources often include customer view, purchase, search, substitution records, as well as product descriptions and hierarchical category information. The tasks of learning/retrieving different product knowledge relations can be naturally brought together under the \textsl{propagation rule} that the substitutable or similar products are more likely to have similar \texttt{complement}, \texttt{co-view}, \texttt{substitute} products as well as descriptions and search terms. This observation motivates us to design the multi-task learning schema leveraging the \textsl{propagation rule}.


Last but not least, we propose to handle the tree-structured hierarchical category by embedding them onto the \textbf{\Poincare ball} \cite{nickel2017poincare}. Several papers including \cite{xie2016representation} have discussed methods for including structured entities in KG embedding. However, they still operate on Euclidean space. While Euclidean space lacks the flexibility to embed tree-structured data properly, some hyperbolic spaces have favorable manifold structures \cite{chamberlain2017neural}. We briefly introduce the technical background for \Poincare embedding in Section \ref{sec:relatedwork}. 

We provide comprehensive comparisons between PKG and ordinary KG in Section \ref{sec:PKGvsKG} and illustrate the critical insight of how \emph{translation} can be carried out via distributed representations in Section \ref{sec:translation}. We introduce our PKG embedding approach in Section \ref{sec:method}. In Section \ref{sec:experiment}, we first demonstrate how KG embedding methods can fail when directly applied to PKG with a real-world e-commerce dataset from \textsl{grocery.walmart.com}. We then show the superior performances of the proposed approach with product knowledge completion tasks and downstream e-commerce tasks. 

\section{Contribution}
\label{sec:contribution}
We conclude several major contributions of this paper as follow.
\begin{itemize}[noitemsep,nolistsep,partopsep=0pt]  
    \item We propose the combination of product relations and PKG for e-commerce with comprehensive illustrations. 
    \item We provide a systematic comparison between PKG and KG.
    \item We thoroughly explain the complex semantics of relations for PKG and show how to deal with the sophisticated relations using distributed representation.
    \item We propose a self-attention-based representation learning model for automatically retrieving relations and learning embeddings for PKG from both user activity data and product information, in an end-to-end fashion.
    \item We demonstrate the meaningfulness and usefulness of the outcome with real-world e-commerce dataset on knowledge completion, search ranking and recommendation tasks.
\end{itemize}

\section{Product Knowledge Graph V.S. Knowledge Graph}
\label{sec:PKGvsKG}
The recent advances in KG embedding, which owes much to several publicly available KG database such as Freebase, DBpedia and YAGO, have led to successful applications in semantic parsing, information extraction, question answering and other NLP tasks \cite{wang2017knowledge}. 
PKG, though important for e-commerce, has received far less attention in existing literature. In this section, we list several critical components of KG embedding and compare them with PKG. 

\textbf{Data source.} KG databases consist of established facts in the triplet form of \textsl{(head entity, relation, tail entity)}. The data source for constructing PKG have multiple modalities, including product catalog information (e.g. description, categories), raw user-product interaction records and others.

\textbf{Model assumption.} The core assumption for KG embedding is that the observed facts in KG database are well-established and plausible. It is not the case for PKG, where observations are much noisier and the facts have not been established.

\textbf{Quantity of relation types.} KG databases often contain thousands of relation types. In PKG, as we discussed before, the major relations can be adequately summarized by \texttt{complement}, \texttt{co-view}, \texttt{substitute}, \texttt{describe}, \texttt{search} and \texttt{IsA}.  

\textbf{Semantics of relation.} Ordinary KG has semantically simple and unambiguous relations such as \texttt{BornIn}, \texttt{DirectorOf},  \texttt{HasWife}. Relations in PKG are semantically more complicated, as we illustrate with the below examples of \texttt{complement} for \texttt{TV}: 
\begin{itemize}
    \item (\texttt{Remote control}, \texttt{complement}, \texttt{TV}): accessory;
    \item (\texttt{TV mount frame}, \texttt{complement}, \texttt{TV}): structural attachment;
    \item (\texttt{Audio speaker}, \texttt{complement}, \texttt{TV}): enhancement;
    \item (\texttt{HDMI Cable switcher}, \texttt{complement}, \texttt{TV}): add-on.
\end{itemize}
In other words, product relations in PKG has much richer semantic meanings since products are designed over a broad range of purposes in the real world. Thus it is impractical to represent the relations merely by translation or/and project operations as their capacity of expressing complex semantic meanings are limited. Distributed representations, on the other hand, is another worthy option. Also, the relations in PKG are all N-to-N, as opposed to the many 1-to-N and 1-to-1 relations in ordinary KG \cite{bordes2013translating}.

\textbf{Additional information.} Both KG and PKG can use entity types, attributes and textual descriptions as additional information to enhance performances, and various methods has been developed for such purposes in KG embedding literature.  

\textbf{Logic rules.} The first order Horn clause , e.g. \texttt{HasWife} $\Rightarrow$ \texttt{Has}-\texttt{Spouse}, motives the logical inference in KG and is exploited by knowledge acquisition and inference. It also helps refine KG embedding \cite{wang2015knowledge}. Unfortunately, relations for PKG can often disobey the Horn clause. On the other hand, PKG do enjoy the \emph{propagation rule} that substitutable products are more likely to have similar relations with other entities.


\textbf{Downstream tasks.}
Most often KG embedding are applied in KG completion tasks \cite{bordes2013translating}. Relation extraction and question answering are two other major directions. As for PKG, product knowledge completion is notably more important due to the \textbf{sparsity} issue in e-commerce data. Relation extraction and question answering can also find their counterparts in e-commerce settings, such as user understanding and searching. A key downstream application for PKG is recommender system. Although several work have proposed KG-enhanced recommendation methods, their application mainly focus on news and movie/book recommendation \cite{wang2018ripplenet,zhang2016collaborative,wang2018dkn}.

\section{Relation Translation via Distributed Representation}
\label{sec:translation}
The distributed representation of words are learnt under the hypothesis that words which appear in a similar context have a similar representation. 
Translation models such as TransE are initially inspired by the linear analogies observed in the distributed representation (word embedding) outcome of \textsl{word2vec} \cite{mikolov2013distributed}, e.g. \textsl{king is to men as queen is to women}. In Euclidean space, the relation of linear analogy corresponds to vector translation: $\mathbf{z}_{\text{king}} - \mathbf{z}_{\text{men}} \approx \mathbf{z}_{\text{queen}} - \mathbf{z}_{\text{women}}$, where $\mathbf{z}_{x}$ denotes the embedding for entity (relation) $x$. The above observations motivate people to represent the relation \texttt{royal} by a translation vector $\mathbf{z}_{\text{royal}}$ such that:
\begin{equation*}
\left\{
\begin{aligned}
&\mathbf{z}_{\text{king}} \approx \mathbf{z}_{\text{men}} + \mathbf{z}_{\text{royal}}  \\
&\mathbf{z}_{\text{queen}} \approx \mathbf{z}_{\text{women}} + \mathbf{z}_{\text{royal}}.
\end{aligned}
\right.
\end{equation*}

Indeed, for ordinary KG where relations are established in fine-grained level, it is convenient and straightforward to define a translation vector (operating on relation-specific spaces) for each relation. However, for PKG where the relations are much more complicated, it is impractical to expect a single translation vector $\mathbf{z}_{\text{complement}}$, even equipped with relation-specific projections, to express various product complementary semantics (e.g. functional complete, structural attachment, enhancement) at the same time. 

The linear analogy observed in distributed representation of words, on the other hand, is capable of expressing complex relation semantics. A recent work proves from a principled manner that \textsl{word2vec} is also learning relational translation in the form of linear analogies \cite{allen2019analogies}. The embeddings are intrinsically optimized to recover the patterns such as \textsl{king is to men as queen is to women}. We state a simplified version of theorem below.

\begin{theorem*}
If "entity $y_1$ is to entity $x_1$ as entity $y_2$ is to $x_2$", then:
\[ 
\mathbf{z}_{y_1} = \mathbf{z}_{x_1} + (\mathbf{z}_{y_2} - \mathbf{z}_{x_2}) + \epsilon 
\]
where $\epsilon$ is the translation error which can depend on the entities and model parameters.
\end{theorem*}

A significant consequence of the theorem is that for each relation (such as \texttt{complement}), the underlying relational semantics, however complicated, can be constructed via entity embeddings that are learned in a similar fashion to \textsl{word2vec}. For instance, instead of explicitly define a translation vector for \texttt{AccessoryTo}, which is a special case of \texttt{complement}, we can expect the model to learn from customer purchase records that: $\mathbf{z}_{\text{AccessoryTo}} \equiv \mathbf{z}_{\text{Xbox}} - \mathbf{z}_{\text{handle}}$ such that 
$\mathbf{z}_{\text{remote control}} + \mathbf{z}_{\text{AccessoryTo}} \approx \mathbf{z}_{\text{TV}}$. We leave the detail of training the entity embeddings to Section \ref{sec:method}.

\section{Related work}
\label{sec:relatedwork}
In this section, we walk through several KG embedding models that we use for building baselines, the background of distributed representation, self-attention and Poincare embedding. To be consistent with original literature we use $\mathbf{r}$ to represent relation embedding.

\textbf{KG embedding models}. Let $\mathbf{h}$ and $\mathbf{t}$ denote head and tail entity of a triplet (\texttt{head entity}, \texttt{relation}, \texttt{tail entity}). \textsl{TransE} uses the distance function for learning from the triplets:
$
d_r(\mathbf{h},\mathbf{t}) = \sum -\|\mathbf{h} + \mathbf{r} - \mathbf{t} \|
$,
where $\|.\|$ can be either $\ell_1$ or $\ell_2$ norm.
\textsl{TransH} further considers relation-specific entity embeddings, i.e $\mathbf{h}_{\bot}^r$ and $\mathbf{t}_{\bot}^r$, defined as entities projected onto the relation-specific hyperplanes. The distance is defined similarly as:
$
d_r(\mathbf{h}, \mathbf{t}) = \sum -\|\mathbf{h}_{\bot}^r + \mathbf{r} -\mathbf{t}_{\bot}^r \|.
$
\textsl{TransR} shares a similar idea to \textsl{TransH} but with the difference that instead of projecting entities onto hyperplanes, \textsl{TransR} considers relation-specific linear subspaces. \textsl{TransD} puts further constraints on the parametrization of relation-specific linear subspaces to achieve higher efficiency.

\textsl{RESCAL} uses the semantic matching distance functions:
$
d_r(\mathbf{h}, \mathbf{t}) = \sum \mathbf{h}^{\intercal} \textbf{M}_r \mathbf{t},
$ where $M_r$ is a matrix associated with relation $r$. \textsl{DistMult} simplifies \textsl{RESCAL} by restricting $M_r$ to be diagonal. The \textsl{Holographic Embedding} (\textsl{HolE}) employs the circular correlation operation $\star$ to combine the advantages of \textsl{RESCAL} and \textsl{DistMult}, where the distance function is defined via $d_r(\mathbf{h}, \mathbf{t}) = \sum \mathbf{r}^{\intercal} (\mathbf{h} \star \mathbf{t})$. 

All the above methods operates on the real space, and to extend the embedding approach of \textsl{DistMult} to complex-valued vectors, \textsl{ComplEx} defines distance as:
$
d_r(\mathbf{h}, \mathbf{t}) = \sum \text{Re}\big(\mathbf{h}^{\intercal} \text{diag}(\mathbf{r}) \mathbf{\bar{\mathbf{t}}}\big),
$ where $\text{Re}(.)$ extracts the real part and $\bar{\mathbf{t}}$ is the conjugate of $\mathbf{t}$.

\textbf{Distributed representation of \textsl{word2vec}}. \textsl{Word2vec} uses the skip-gram model to learn the distributed representation of words. The score function is defined as the total log-probability of observing the words given their contexts: $
S = \sum_{i} \sum_{j \in \text{Context}(i,c)}\log p(e_i | e_j),
$
where $\text{Context}(i,c)$ is the set of neighbours of entity $i$ within a window of size $c$. Each probability term is computed with softmax function such that
\begin{equation}
\label{eqn:word2vec}
p(e_i | e_j) = \frac{\exp \big((\zbf_i^{O})^{\intercal}(\zbf_i^{I})\big)}{\sum_k \exp \big((\zbf_k^{O})^{\intercal}(\zbf_i^{I})\big)},
\end{equation}
where $\Zbf^I$ and $\Zbf^O$ are the "input" and "output" word embeddings. To avoid computing the summation term over the whole vocabulary, hierarchical softmax and negative sampling are often employed as approximation methods for the computational efficiency \cite{mikolov2013distributed}. 

\textbf{Self-attention}. The key idea behind the attention mechanism is that only part of the input sequence is relevant to the output, and the model should pay more attention on the relevant part. As an add-on component, attention mechanism has been widely applied to image captioning and machine translation. Recently, the purely attention-based sequence-to-sequence \textsl{Transformer} model achieves state-of-the-art performances in machine translation tasks \cite{vaswani2017attention}. After discarding the RNN structures for modelling sequences, \textsl{Transformer} relies heavily on the self-attention module. The authors also demonstrate the improved model interpretation and computation efficiency for the \textsl{Transformer} \cite{vaswani2017attention}.

\textbf{\Poincare embedding}. Embedding entities in the Euclidean vector spaces does not account for their latent hierarchical structures. Hyperbolic space such as the \Poincare Ball, on the other hand, can represent hierarchy structures and similarity more parsimoniously. We refer the readers to \cite{chamberlain2017neural} for a comprehensive introduction on hyperbolic embedding. The main difference induced by operating on Poincare Ball lies in the distance metric, which is defined as:
\begin{equation}
\label{eqn:poincare}
    d_{\text{\Poincare}}(e_i, e_j) = \text{arcosh}\big(1+2 \frac{\|\zbf_i - \zbf_j\|^2_2}{(1-\|\zbf_i\|_2^2)(1-\|\zbf_j\|_2^2)} \big).
\end{equation}
Loss functions developed upon (\ref{eqn:poincare}) can be optimized via Riemannian optimization methods according to the Riemannian structure of \Poincare Ball. Efficient implementation of stochastic gradient descent on Riemannian manifolds has also been developed \cite{bonnabel2013stochastic}.

\section{Methodology}
\label{sec:method}
When learning PKG embedding according to the proposed approach, we use a generic e-commerce dataset that consists of session-based purchase and view sequences, product substitution records, the search-and-click records, product descriptions and hierarchical category labels. We point out that the proposed approach is still applicable when one or more of the above data modalities are unavailable. The notations are summarized in Table \ref{tab:notation}. 

\begin{figure}
    \centering
    \includegraphics[scale=0.27]{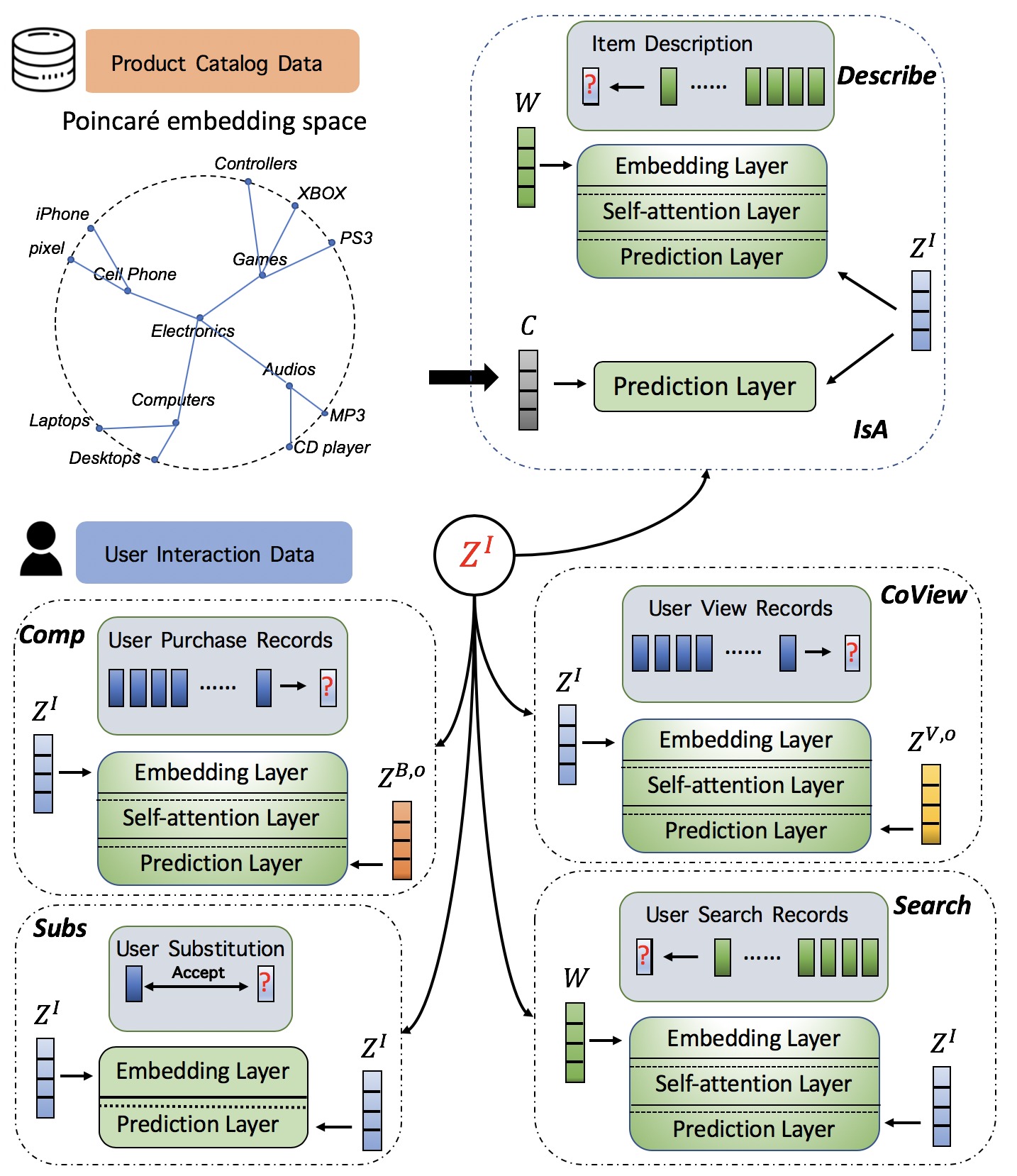}
    \caption{Architecture for learning PKG embedding.}
    \label{fig:architect}
\end{figure}

\begin{table}[thb]
    \centering
    \begin{tabular}{L{1.7cm}|L{6cm}}
    \hline 
        Notation & Description \\ \hline
        
        $d \in \mathbb{N}$ & entity embedding dimension \\ \hline 
        
        $l \in \mathbb{N}$ & the maximum sequence length for predicting target entity \\ \hline
        
        $\mathcal{I}$,$\mathcal{W}$,$\mathcal{C}$ & item (product), word and category label set \\ \hline 
        
        
        $\mathcal{B}$, $\mathcal{V}$, $\mathcal{S}$ & customer session-based co-buy, co-view and substitution acceptance records, for instance, $\mathcal{B}_i = (\mathcal{I}_{i_1}, \mathcal{I}_{i_2}, \ldots, \mathcal{I}_{|\mathcal{B}_i|})$, $\mathcal{S}_j = (\mathcal{I}_{j_1}, \mathcal{I}_{j_2})$ \\ \hline 
        $\mathcal{Q}$, $\mathcal{D}$, $\mathcal{L}$ & query (search), description, category labels for products, i.e. $\mathcal{Q}_{\mathcal{I}_j} \subset \mathcal{W}$, $\mathcal{D}_{\mathcal{I}_j} \subset \mathcal{W}$, $\mathcal{L}_{\mathcal{I}_j} \subset \mathcal{C}$  \\ \hline
        
        $\Zbf^I \in \mathbb{R}^{|\mathcal{I}| \times d}$ & product "input" embedding similar to that of \textsl{word2vec}, e.g. $\Zbf^I_{\mathcal{I}_j}$ denotes product "input" embedding for item $\mathcal{I}_j$ \\ \hline
        
        $\Zbf^{B,O}, \Zbf^{V,O} \in \mathbb{R}^{|\mathcal{I}| \times d}$ & product "output" embeddings for modelling co-buy and co-view records  \\ \hline
        
        $\Wbf \in \mathbb{R}^{|\mathcal{W}| \times d}$ & word entity embeddings \\ \hline 
        
        $\Cbf \in \mathbb{R}^{|\mathcal{C}| \times d}$ & category entity embeddings \\ \hline
        
        $\mathbf{P} \in \mathbb{R}^{l \times d}$ & positional encoding matrix for self-attention \\ \hline
    \end{tabular}
    \caption{Notation}
    \label{tab:notation}
\end{table}

\subsection{Modelling \texttt{substitute} Relation}
\label{sec:substitution}
According to the \emph{propagation rule} mentioned in Section \ref{sec:introduction} and \ref{sec:PKGvsKG}, the \texttt{substitute} relation is the key to bring together the various relations for PKG embedding. Recall the distributional hypothesis which states that the contextually similar words have similar representations. The \emph{propagation rule} implies, and it is indeed observed in \textsl{word2vec} outcome, that similar words lay closer to each other in the word "input" embedding space. In analogy, substitutable products are also expected to have similar product "input" embeddings. When the customer substitution records are available, we can directly model the product "input" embeddings $\Zbf^I$. Since the \texttt{substitute} relation is symmetric, which means if A can substitute B then B can also substitute A, for each substitution accepted by customers we define the substitution score as:
\begin{equation}
\label{eqn:subs}
    S_{\text{sub}} = \sum_{(e_1,e_2) \in \mathcal{S}}\log p(e_1,e_2) \equiv \sum_{(e_1,e_2) \in \mathcal{S}} \log \frac{\exp\big((\Zbf^I_{e_1})^{\intercal} \Zbf^I_{e_2}\big)}{\sum_{i \in \mathcal{I}} \exp\big((\Zbf^I_{i})^{\intercal} \Zbf^I_{e_2}\big)}.
\end{equation}
We shall see in the next sections that $\Zbf^I$ builds the bridge between modelling various relations by exploiting the \emph{propagation rule}.

\subsection{Self-attention Mechanism for \texttt{complement}, \texttt{co-view}, \texttt{describe} and \texttt{search} Relations}
\label{sec:self-attention}
To extract \texttt{complement}, \texttt{co-view}, \texttt{search} and \texttt{describe} relations from the noisy customer purchase, view, search activities and product descriptions respectively, we use self-attention module with problem-specific modifications. For notation simplicity, in this part (Section \ref{sec:self-attention}) we use $\Zbf^O$ to denote product "output" embedding as a whole. Also, the sequence length $l$ may vary across tasks.

\textbf{Embedding layer for self-attention.} The embedding layer takes an ordered sequence of entities (products or words) as input. To model positional information, self-attention uses positional encoding such that each position $k$ maps to a vector $\mathbf{P}_k \in \mathbb{R}^d$. The entity sequence is truncated at the maximum length $l$, which we denote by $\ebf=(e_1, \ldots, e_l)$. The embedding layer takes the sum of entity embeddings (either $\Zbf^I$ or $\Zbf^O$) and their corresponding position encoding from $\mathbf{P}$, and the output is given by:
\[
\mathbf{E}^I_{\ebf} = [\Zbf^I_{e_1} + \mathbf{P}_1, \ldots, \Zbf^I_{e_1} + \mathbf{P}_l]^{\intercal} \in \mathbb{R}^{l \times d}
\]
and $
\mathbf{E}^O_{\ebf} = [\Zbf^O_{e_1} + \mathbf{P}_1, \ldots, \Zbf^O_{e_1} + \mathbf{P}_l]^{\intercal} \in \mathbb{R}^{l \times d}$.

\textbf{Self-attention layer.} We use the scaled dot-product attention as building block, which is defined as:
\begin{equation}
    \text{Attn}(\Qbf, \Kbf, \Vbf) = \text{softmax}\Big(\frac{\Qbf \Kbf^{\intercal}}{\sqrt{d}} \Big)\Vbf,
\end{equation}
where $\Qbf$ represents the "queries", $\Kbf$ the "keys" and $\Vbf$ the "values". Since each row of $\Qbf, \Kbf, \Vbf$ correspond to an entity, the dot-product attention layers outputs the weighted sum of the entity embeddings in $\Vbf$, where the weights reflects the pairwise "query-key" interactions in the entities sequence.

Given our distributed representation setting, it is intuitive to consider using $\mathbf{E}^I$ as the "queries" and $\mathbf{E}^O$ as the "keys" since they embed the contextual and positional relatedness information of the entity pairs. 
We also choose to use $\mathbf{E}^I$, the entity "input" embedding with positional encoding, as the "values" $\mathbf{V}$, for reasons which we will later explain.
So the output of our attention layer is given by:
$
    \Hbf = \text{Attn}(\mathbf{E}^I, \mathbf{E}^O, \mathbf{E}^{I}).
$
To see how self-attention layer is capable of assigning weights to entities such that the output only focus on related part of the sequence, we point out that $\Hbf$ can be alternatively expressed as:
\[
    \Hbf_i = \sum_{j} \alpha_{ij} \times \mathbf{E}^{I}_j,
\]
with the weights $\alpha_{ij} \equiv (\mathbf{E}^I_i) ^{\intercal} \mathbf{E}^O_i$ capturing the contextual and positional relation between entity $e_i$ and $e_j$.

However, taking direct inner-product between $\mathbf{E}^I$ and $\mathbf{E}^O$ does not take account of the interactions between different latent dimensions. It can impair the expressiveness of the self-attention layer. Instead, we add a two-layer point-wise feed-forward network to the entity "input" and "output" embeddings before passing them to the dot-product attention, i.e. 
\begin{equation}
\begin{split}
    & \text{FFN}(\mathbf{E}_i) \equiv \text{ReLU}(\mathbf{E}_i \mathbf{\Theta}_1+\mathbf{b}_1) \mathbf{\Theta}_2+\mathbf{b}_2, \\
    &\mathbf{F}^I_i = \text{FFN}(\mathbf{E}^I_i), \  \mathbf{F}^O_i = \text{FFN}(\mathbf{E}^O_i), \ \forall i \in \{1,\ldots,l\}.
\end{split}
\end{equation}
where $\mathbf{\Theta}_1, \mathbf{\Theta}_2 \in \mathbb{R}^{d\times d}$ are parameter matrices and $\mathbf{b}_1, \mathbf{b}_2 \in \mathbb{R}^d$ are bias terms. So the complete self-attention layer is given by:
\begin{equation}
     \Hbf = \text{Attn}(\ebf) \equiv \text{Attn}(\mathbf{F}^I_{\ebf}, \mathbf{F}^O_{\ebf}, \mathbf{E}^{I}_{\ebf}), \Hbf \in \mathbb{R}^{l \times d}.
\end{equation}


\textbf{Prediction layer.} When modelling with customer purchase and view data, it is straightforward to use the previously purchased/viewed product sequences to predict next purchase/view product. Similar to that of \textsl{word2vec} in (\ref{eqn:word2vec}), we  estimate the \\ $p(e_{l+1} | e_1, \ldots, e_l)$ where $e$ can be either word token or product. 

To compute $p(e_{l+1} | e_1, \ldots, e_l)$, we need an aggregated representation of the context sequence $\ebf = (e_1, \ldots, e_l)$. By plugging in our self-attention layer as sequence aggregator, we obtain:
\begin{equation}
\label{eqn:attn-word2vec1}
\begin{split}
    &\log p(e_{l+1} | \ebf) = (\Zbf^O_{e_{l+1}})^{\intercal} \text{Attn}(\ebf) - \log \sum_{i \in \mathcal{I}} \exp\big((\Zbf^O_{i})^{\intercal} \text{Attn}(\ebf) \big) \\ 
    &=(\Zbf^O_{e_{l+1}})^{\intercal} \sum_{j=1}^l \alpha_{ij} \Zbf^I_{e_j} - \log \sum_{i \in \mathcal{I}}  \exp\big((\Zbf^O_{i})^{\intercal} \sum_{j=1}^l \alpha_{ij} \Zbf^I_{e_j} \big) + C,
\end{split}
\end{equation}
where $C$ is some constant term unrelated to learning entity embeddings. We point out that (\ref{eqn:attn-word2vec1}) is an extension of (\ref{eqn:word2vec}) where the context sequence is aggregated by the attention weights that are part of our self-attention layer. The inner product terms in (\ref{eqn:attn-word2vec1}) also explains why we use the entity "input" embedding $\Zbf^I$ as the "value" matrix in the dot-product attention. By doing so, we are still modelling with the inner products between the product "input" and "output" embeddings ($\Zbf^I$ and $\Zbf^O$).

The description and search data can be modelled with the same prediction setting. Given the desription/search words, we predict the target product. Since $\Zbf^I$ carries the substitutable product information, under the \emph{propagation rule} products that are closer in terms of $\Zbf^I$ should have similar descriptions and search terms. Therefore, we use the word embedding to predict product "input" embedding:
\begin{equation}
\small 
\label{eqn:attn-word2vec2}
\begin{split}
\log p(e_{l+1} | \ebf) &= (\Zbf^I_{e_{l+1}})^{\intercal} \sum_{j=1}^l \alpha_{ij} \Wbf^I_{e_j} - \log \sum_{i \in \mathcal{I}}  \exp\big((\Zbf^I_{i})^{\intercal} \sum_{j=1}^l \alpha_{ij} \Wbf^I_{e_j} \big) + C,
\end{split}
\end{equation}
where the entity sequence $\ebf$ is from search or description.

\textbf{Score functions.} With the outputs from the prediction layer, the score functions can be given in a straightforward manner, e.g.  
\begin{equation}
\begin{split}
    S_{\text{complement}} = \sum_{(e_{l+1}, \ebf) \in \mathcal{B}} \log p(e_{l+1} | \ebf).
\end{split}
\end{equation}
The score functions of $S_{\text{co-view}}$, $S_{\text{describe}}$, $S_{\text{search}}$ are also computed according to the data of $\mathcal{V}$, $\mathcal{D}$, $\mathcal{S}$ in the same fashion. We do not write them down to avoid unnecessary repetitions.

\subsection{\Poincare Embedding for Category Hierarchy and \texttt{IsA} Relation}
\label{sec:poincare}
To learn the distributed representations of the hierarchical categories, which are originally tree-structured symbols, we employ the \Poincare embedding for the best practice. When a category symbol $c_1$ is a child node of $c_2$, their distance in terms of $d_{\text{\Poincare}}$ is supposed to be small. To be consistent with the score functions for learning other relations, we also estimate $p(c_1|c_2)$ as a classification task with softmax function:
\begin{equation}
\label{eqn:label}
    p(c_1|c_2) = \frac{\exp\big( -d_{\text{\Poincare}}(c_1,c_2) \big)}{\sum_{c \in \mathcal{C}} \exp \big(-d_{\text{\Poincare}}(c,c_2)\big)}.
\end{equation}
Notice that the category label embddings $\Cbf$ are independent from the product and word embeddings in other tasks. Therefore, it is convenient to pre-train the \Poincare embeddings for the category labels and then pass them to the final task of modelling \texttt{IsA} relation. 

Now that the category embeddings $\Cbf$ are fixed, the \texttt{IsA} relation can be directly formulated as a multi-class classification problem, where we use product embedding to predict their category labels. Again we use the product "input" embedding $\Zbf^I$ for the same reason as using $\Zbf^I$ for modelling \texttt{describe} and \texttt{search}. For consistency we also use softmax function in the score function $S_{\text{IsA}}$:
\begin{equation}
\label{eqn:isa}
    S_{\text{IsA}} = \sum_{i \in |\mathcal{I}|} \sum_{j \in \mathcal{L}_{i}} \log \frac{\exp\big( \Cbf_j^{\intercal} \Zbf^I_i \big)}{\sum_{c \in \mathcal{C}}\exp\big( \Cbf_c^{\intercal} \Zbf^I_i \big)}.
\end{equation}
We point out that the (log-)normalization terms in (\ref{eqn:subs}), (\ref{eqn:attn-word2vec1}), (\ref{eqn:attn-word2vec2}), (\ref{eqn:label}) and (\ref{eqn:isa}) can all be efficiently approximated by negative sampling.

\subsection{Multitask Training}
\label{sec:multitask}

Although the score functions associated with each task have simple forms and unequivocal interpretations, it is unclear how to combine them into an overall score for optimal results. Despite that many methods have been proposed for multi-task learning with deep neural networks such as \textsl{GradNorm} \cite{chen2017gradnorm}, \textsl{MGDA} \cite{sener2018multi} and \textsl{uncertainty weighting} \cite{kendall2018multi}, they assume a shared network structure across tasks. In our setting, however, only the product "input" embedding is shared in all tasks. Also, most of the above approaches require computing all the gradients during each update, which is infeasible in our case due to the vast number of free parameters in our method. Similarly, the various searching algorithms for detecting the best-weighted combination of individual score functions \cite{zitzler1998evolutionary} are also computationally impractical due to the large scale of real-world e-commmerce data. Interestingly, a recent work observes that there is no clear consensus on correctly training multi-task model in an NLP setting similar to ours \cite{subramanian2018learning}. 

As a compromise, we choose the simple yet effective training method described in \cite{sanh2019hierarchical}. After each training epoch, a task is randomly selected, and a batch of dataset associated with this task is sampled for training. The sampling is done in a weighted fashion, so the probability of sampling a task is proportional to the relative size of each dataset. The sampling-then-training process is repeated until the metrics for each task stop improving on validation dataset.

\subsection{Prediction for Downstream Tasks}
\label{sec:prediction}

The downstream PKG tasks, such as knowledge completion, search ranking and recommendation, require prediction with the learned PKG embeddings. Task details are discussed in Section \ref{sec:tasks}. When predicting or ranking the candidate tail entities $t$ given head entity $h$ (for \texttt{complement}, \texttt{co-view}, \texttt{substitute}) or entities $\mathbf{h}$ (for \texttt{search}, \texttt{describe}, recommendation), the objective is the $p(t | h)$ or $p(t | \mathbf{h})$ which we can compute according to Table \ref{tab:prediction}. 

\begin{table}[]
    \centering
    \begin{tabular}{l|c}
    \hline 
        Task & $p(t | h)$ / $p(t | \mathbf{h})$ \\ \hline 
        \texttt{substitute} & $\propto \exp \big( (\Zbf^I_t)^{\intercal}\Zbf^I_h \big)$, $h,t \in \mathcal{I}$ \\
        \texttt{complement} & $\propto \exp \big( (\Zbf^{B,O}_t)^{\intercal}\Zbf^I_h \big)$, $h,t \in \mathcal{I}$ \\
        \texttt{co-view} & $\propto \exp \big( (\Zbf^{V,O}_t)^{\intercal}\Zbf^I_h \big)$, $h,t \in \mathcal{I}$ \\
        \texttt{search} & \multirow{ 2}{*}{$\propto \exp \big( (\Zbf^{I}_t)^{\intercal}\text{Attn}(\mathbf{h}) \big)$, $\mathbf{h} \subset \mathcal{W},t \in \mathcal{I}$} \\
        \texttt{describe} & \\
        \texttt{Isa} & $\propto \exp \big( (\Zbf^{I}_t)^{\intercal}\mathbf{C}^I_h \big)$, $h \in \mathcal{C},t \in \mathcal{I}$ \\
        recommend & $\propto \exp \big( (\Zbf^{B,O}_t + \Zbf^{V,O}_t)^{\intercal}\text{Attn}(\mathbf{h}) \big)$, $\mathbf{h} \subset \mathcal{I},t \in \mathcal{I}$ \\ \hline
    \end{tabular}
    \caption{Prediction for each task (relation) according to trained embeddings.}
    \label{tab:prediction}
\end{table}

\section{Experiment and Result}
\label{sec:experiment}
We design the experiments to answer the following questions:

\textbf{Q1}: Is the proposed multi-task learning schema reasonable? 

\textbf{Q2}: Other than knowledge completion, how does the PKG embedding benefit downstream e-commerce tasks? 


\textbf{Q3}: Why KG embedding methods fail to work when directly applied to raw e-comerce dataset?

\textbf{Q4}: If a product relation graph is available for KG embedding methods, can the proposed approach still outperform the baselines?

\subsection{Dataset}
\label{sec:dataset}
We evaluate our approach on a real-world e-commerce dataset obtained from \textsl{grocery.walmart.com}, the largest online grocery shopping platform in the U.S. Product catalog information of the dataset are summarized in Table \ref{tab:dataset}.

\begin{table}[htb]
    \centering
    \begin{tabular}{|L{1.5cm}|L{6cm}|}
    \hline
        products & The dataset contains $\sim$140,000 common grocery products covering a broad range from food to appliances \\ \hline
        description & Each product is provided with a short description (containing name and brand) as shown on the website. Usually the descriptions have 20 - 100 words. \\ \hline
        category hierarchy & Each product is assigned to a cateogry hierarchy in the form of \{subcategory, category, department, super-department\}, and there are 1,198 subcategories, 228 categories, 28 departments and 9 super-departments. \\ \hline
    \end{tabular}
    \caption{Summary of product catalog data.}
    \label{tab:dataset}
\end{table}

\textbf{Session data}. We are provided with $\sim$40 million session records with the views, purchases, search queries and the products that are clicked according to the search queries. 

\textbf{Substitution data}. When products went out of stock, substitutions are recommended for the customers where they can choose to accept or deny. The dataset consists of the accepted substitutions. Around 70,000 products have been substituted.

\textbf{Preprocess}. We remove products that have less than ten total appearances (purchase, view, searched, substitution), which leaves us $\sim$100,000 products. Words with less than three appearances (description, search query) are removed. We point out that our approach is capable of handling \textbf{sparsity} issue, but to make sure that the baselines can work properly, we filter out infrequent entities.

\textbf{Product relation graph (PRG)}. We are able to construct the product relation graph for \texttt{complement}, \texttt{co-view} and \texttt{substitute}. We first build a weighted graph such that the edge between node A and B is the number of session that these two products have been co-viewed, co-purchased or substituted. For instance, we use $\Xbf$ to denote the resulting adjacency matrix for co-purchase, so 
\[
\Abf_{ij} = \# \{k | (\mathcal{I}_i, \mathcal{I}_j) \subset \mathcal{B}_k \}.
\]
The normalized adjacency matrix can be computed via 
\[
\tilde{\Abf} = \Dbf^{-\frac{1}{2}}\Abf \Dbf^{-\frac{1}{2}},
\]
where $\Dbf$ is a diagonal matrix with $\Dbf_{ii}=\sum_j \Abf_{ij}$. We then train a \textsl{biased random walk} with the normalized adjacency matrix and obtain top-K related neighbors for each product. After initial inspections we find that $K=20$ gives reasonable results for most products. The parameters are then tuned such that the top-20 related products achieve best link prediction performance on a hold-out subset in terms of \emph{hitting rate}. Then we treat the top-20 related products for each product as known facts and obtain the product relation graph $\Gbf_{\text{buy}}$. $\Gbf_{\text{view}}$ and $\Gbf_{\text{subs}}$ can be constructed in the same way.


\subsection{Baseline Methods}
\label{sec:baseline}
To answer \textbf{Q3} and \textbf{Q4}, we implement classic KG embedding methods on both raw data and data enhanced with $\Gbf_{\text{buy}}$, $\Gbf_{\text{view}}$ and $\Gbf_{\text{subs}}$. When learning KG embedding from raw data, we directly enumerate all triplets from product catalog information, session data and substitution data, e.g. 
\begin{equation*}
\begin{split}
    \Xbf_{\text{view}} = \{(\mathcal{I}_1, \texttt{co-view}, \mathcal{I}_2)\ | \ (\mathcal{I}_1, \mathcal{I}_2) \subset \mathcal{V}_i, 1 \leq i \leq |\mathcal{V}| ) \} \\
    \Xbf_{\text{sub}} = \{(\mathcal{I}_1, \texttt{substitute}, \mathcal{I}_2)\ | \ (\mathcal{I}_1, \mathcal{I}_2) \in \mathcal{S}_i, 1 \leq i \leq |\mathcal{S}| ) \}.
\end{split}
\end{equation*}
$\Xbf_{\text{buy}}$, $\Xbf_{\text{describe}}$, $\Xbf_{\text{search}}$ and $\Xbf_{\text{IsA}}$ are constructed in the same way.

Although the above construction mechanism creates a massive number of triplets, it is not clear how to effectively implement downsampling. Therefore, we only implement \textbf{TransE}, \textbf{TransD} and \textbf{DistMult} on this generated dataset (\textbf{No PRG}). 

With the pre-trained product relation graph $\Gbf_{\text{buy}}$, $\Gbf_{\text{view}}$, $\Gbf_{\text{subs}}$ replacing $\Xbf_{\text{buy}}$, $\Xbf_{\text{view}}$ and $\Xbf_{\text{sub}}$, the number of triplets is considerably decreased in the enhanced dataset (\textbf{With PRG}). Therefore we train \textbf{TransE}, \textbf{TransH}, \textbf{TransR}, \textbf{TransD}, \textbf{RESCAL}, \textbf{DistMult} and \textbf{ComplEx} for comprehensive comparisons. 

Since the KG embedding methods are not designed for recommendation, we further include \textbf{Factorization Machine} (\textbf{FM}) \cite{rendle2010factorization}, \textbf{Bayesian Personalized Ranking} (\textbf{BPR}) \cite{rendle2009bpr}, \textbf{Prod2vec} \cite{vasile2016meta} and \textbf{Triple2vec} \cite{wan2018representing} as baselines. We choose these methods among others because they also learn latent representations of products.

\subsection{Implementation Details}
\label{sec:implementation}
To be consistent with the original implementation, we also use stochastic gradient descent for the KG embedding models with learning rate selected among \{0.001, 0.005, 0.01, 0.1\}. The margin $\gamma$ for the translation models are selected among \{1, 2, 5, 10\}. We also choose the $\ell_1$ or $\ell_2$ norm according to validation performance measured by \emph{hitting rate} on the knowledge completion task that we describe in Section \ref{sec:tasks}. All the above models are trained with negative sampling, where for each positive triplet we sample \emph{three} negative (corrupted) triplets. All the implementations are in \textsl{Tensorflow}, where we use the open-source framework for knowledge embedding (\textsl{OpenKE}) as reference for the KG embedding baselines.

For the proposed approach, we also use stochastic gradient optimizer with learning rate selected among \{0.001, 0.005, 0.01, 0.1\}. After initial data analysis, we choose $l_{\text{buy}}=20$, $l_{\text{view}}=50$, $l_{\text{describe}}=200$, $l_{\text{search}}=10$ as the maximum sequence lengths for the self-attention mechanism in corresponding tasks. The \Poincare embedding for category hierarchy is pre-trained with the default setting proposed in \cite{nickel2017poincare}. We also use three negative samples for our approach in the same manner as that of \textsl{word2vec}. 

Finally, we set the dimension of embeddings and latent factors to 100 for all methods. The recommendation baseline methods are also tuned for best performance on \emph{hitting rate} in validation data.

\subsection{Tasks and Evaluation Metrics}
\label{sec:tasks}
\textbf{Knowledge completion}. We focus on the link prediction task for knowledge completion and entity classification. While knowledge completion examines the \textbf{usefulness} or PKG embedding, entity classification examines their \textbf{meaningfulness}.  For knowledge completion, we evaluate the prediction of tail entity when given head entity and relation. The relation is one of \{\texttt{complement}, \texttt{co-view}, \texttt{substitute}\}.  We use top-10 hitting rate (\textbf{HIT@10}) and normalized discounted cumulative gain (\textbf{NDCG@10}) to evaluate the candidate rankings. For entity classification, we predict the \textsl{category} and \textsl{department} labels for products, using multi-class logistic regression with product "input" embedding as features. We report the \textbf{micro-F1} and \textbf{macro-F1} scores for the classification outcome.

\textbf{Search ranking}. We rank all products according to given search query and compute the top-10 recall (\textbf{R@10}) and mean average precision (\textbf{MAP@10}).

\textbf{Recommendation}. Since we only have session-based customer purchase records, we examine recommendation performance using with-in basket recommendation, i.e. given what the customer has purchases so far in the current and previous sessions, we predict the next impression (purchase or view) in the same session. We also report \textbf{HIT@10} and \textbf{NDCG@10}.

\begin{table*}[bht]
\footnotesize
    \centering
    \begin{tabular}{c|cccccc|cccc}
    \hline
        Task & \multicolumn{6}{c}{Link prediction} & \multicolumn{4}{c}{Product classification} \\ \hline
        Relation & \multicolumn{2}{c}{\texttt{complement}} & \multicolumn{2}{c}{\texttt{co-view}} & \multicolumn{2}{c}{\texttt{substitute}} & \multicolumn{2}{c}{\texttt{IsA} (\textsl{category})} & \multicolumn{2}{c}{\texttt{IsA} (\textsl{department})}  \\ 
        Metric & \textbf{Hit@10} & \textbf{NDCG@10} & \textbf{Hit@10} & \textbf{NDCG@10} & \textbf{Hit@10} & \textbf{NDCG@10} & \textbf{micro-F1} & \textbf{macro-F1} & \textbf{micro-F1} & \textbf{macro-F1}\\ 
        & (a1) & (a2) & (a3) & (a4) & (a5) & (a6) & (a7) & (a8) & (a9) & (a10) \\ 
        \hline
        \textbf{TransE} (No PRG) & 1.84  & 1.06  & 3.27  & 2.11  & 12.04  &   6.56 & 42.33  & 69.44  & 51.72  & 76.53 \\
        \textbf{TransD} (No PRG) & 1.97  & 1.08  &  3.51 & 2.24  &  13.69 &  6.90  & 41.82  & 67.65  & 50.29 & 75.45  \\
        \textbf{DistMult} (No PRG) & 3.47  & 1.88  &  6.58 & 3.41  &  20.64 & 9.96 &  53.75 & 74.69  & 61.42  & 80.73  \\
        \textbf{TransE} (With PRG) & 3.65  & 1.82  &  6.95  & 3.90  & 30.22 &  13.41 & 45.43  & 74.93  & 55.89 & 81.81  \\
        \textbf{TransH} (With PRG) & 4.13  & 1.79  & 6.88  & 2.89  & 30.37  & 13.56 & 41.94  &  64.09  & 50.12  & 72.95  \\
        \textbf{TransR} (With PRG) & 6.06  & 2.35  &  8.17 & 3.43  & 31.25  & 14.88 & 46.37  & 72.74  & 53.95  &  74.11 \\
        \textbf{TransD} (With PRG) & 4.26  & 1.95  & 7.03  & 2.97  &  20.71 & 9.86 &  50.36 & 71.02  & 59.62  &  82.43 \\
        \textbf{RESCAL} (With PRG) & 1.64  & 0.97  & 1.63  & 0.87  & 12.46  & 5.76 & 62.89  & 86.27  & 72.27  & 90.97  \\
        \textbf{DistMult} (With PRG) & 5.69  & 2.47  & 9.64  & 4.05  & 30.64  & 12.25 & \underline{\textsl{68.25}}  & \underline{\textsl{94.23}}  & 72.09  &  92.94 \\
        \textbf{ComplEx} (With PRG) & \underline{\textsl{7.81}}  & \underline{\textsl{3.36}}  &  \underline{\textsl{12.38}} & \underline{\textsl{5.77}}  & \underline{\textsl{31.25}}  & \underline{\textsl{12.60}}  &  67.46 & 94.02  & \underline{\textsl{72.54}}  & \underline{\textsl{97.71}}  \\
        \textbf{Our approach} & \textsl{\textbf{14.53}}  & \textsl{\textbf{7.67}}  & \textsl{\textbf{20.84}}  & \textsl{\textbf{10.26}}  & \textsl{\textbf{34.58}}  &  \textsl{\textbf{14.77}} & \textsl{\textbf{68.62}} & \textsl{\textbf{95.17}} & \textsl{\textbf{74.61}} & \textsl{\textbf{99.60}} \\ \hline
    \end{tabular}
    \caption{Testing performances on the knowledge completion tasks. The results are average over three runs and reported in \%. The best performing method in each row is boldfaced, and the second best method in each row is underlined. The labels beneath the metric name corresponds to the labels in Figure \ref{fig:corr_and_radar}.}
    \label{tab:result-knowledge-completion}
\end{table*}

\subsection{Training and Testing}
\label{sec:training-testing}

The train-validation-test split is more involved because we are experimenting on several tasks with various baselines. Since the user activity data are timestamped, we split $\Xbf_{\text{view}}$, $\Xbf_{\text{buy}}$, $\Xbf_{\text{sub}}$, $\Xbf_{\text{search}}$ into 80\%-10\%-10\% according to their chronological order. The product relation graphs are trained on $\Xbf_{\text{view}}^{\text{train}}$, $\Xbf_{\text{buy}}^{\text{train}}$, $\Xbf_{\text{sub}}^{\text{train}}$ and validated on $\Xbf_{\text{view}}^{\text{validate}}$, $\Xbf_{\text{buy}}^{\text{test}}$ $\Xbf_{\text{sub}}^{\text{validate}}$. 

\textbf{Knowledge completion}. We do the train-validation-test split on product relation graphs $\Gbf_{\text{buy}}$, $\Gbf_{\text{view}}$ and $\Gbf_{\text{sub}}$ into 80\%-10\%-10\% with one condition that there is no isolated node in training graph. The proposed approach and baseline KG embedding methods are tested on $\Gbf_{\text{buy}}^{\text{test}}$, $\Gbf_{\text{view}}^{\text{test}}$ and $\Gbf_{\text{sub}}^{\text{test}}$ for \texttt{complement}, \texttt{co-view} and \texttt{substitute}. We also randomly select 10\% of the products, mask all of their category information during training and predict their \textsl{category} and \textsl{department} in testing. For fair comparisons,  when training our model, we remove all the related data from $\Xbf_{\text{view}}$, $\Xbf_{\text{buy}}$, $\Xbf_{\text{sub}}$ and $\Xbf_{\text{describe}}$ that can cause information leak. 

\textbf{Search ranking}. We test the performances on search ranking with $\Xbf_{\text{search}}^{\text{test}}$, on queries that have appeared in training dataset as well as new queries. For baseline KG embeddings methods, we take the average of entity word embeddings as query embedding. 

\textbf{Recommendation}. For all the baselines considered, the within-session recommendation is trained on $\Xbf_{\text{buy}}^{\text{train}} \cup \Xbf_{\text{view}}^{\text{train}}$ and tested on $\Xbf_{\text{buy}}^{\text{test}} \cup \Xbf_{\text{view}}^{\text{test}}$.

\subsection{Analysis on Multi-task Training}
\label{sec:multi-task}
\begin{figure}
    \centering
    \includegraphics[width=0.45\textwidth]{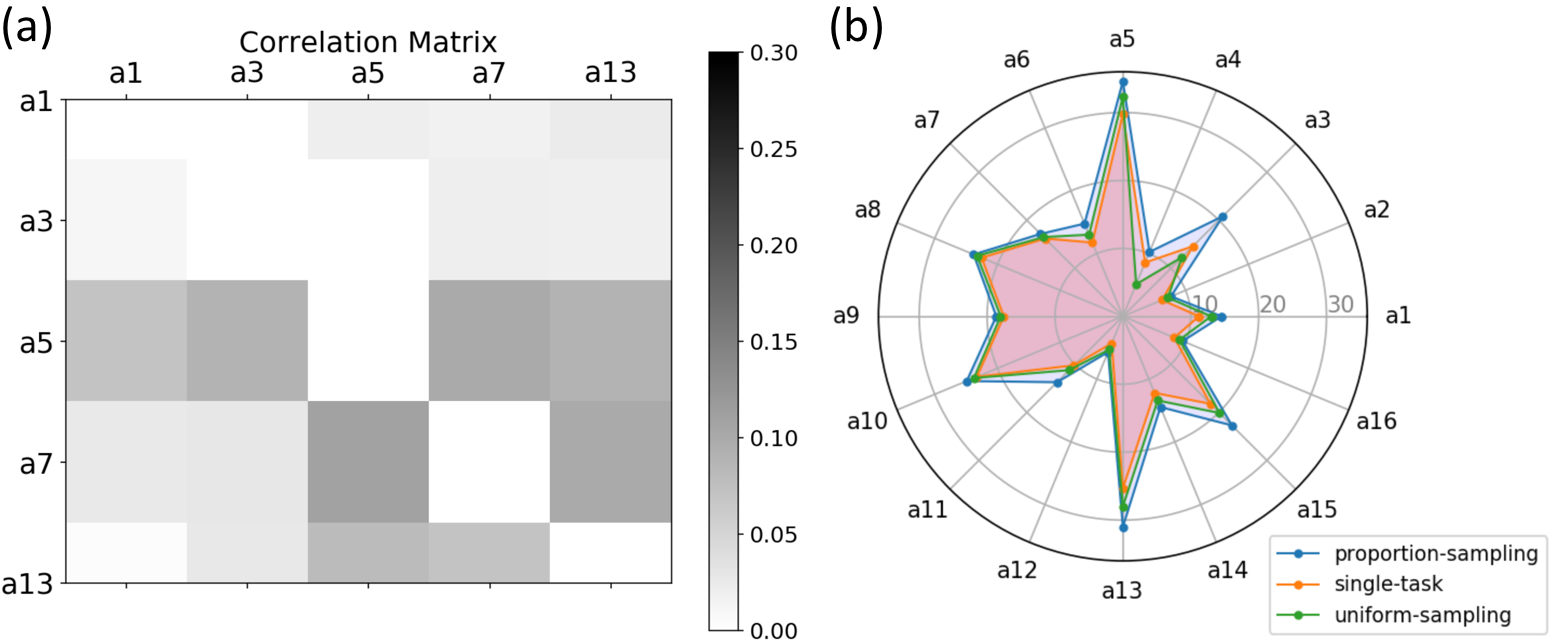}
    \caption{\small (a) Heatmap for the task correlations $\rho$ (b) Test performances of different training schedules on all the tasks. The symbols for each task can be found beneath the metrics in Table \ref{tab:result-knowledge-completion} and \ref{tab:result-search}. Micro-F1 and macro-F1 are divided by 4 for presentation purpose.}
    \label{fig:corr_and_radar}
\end{figure}

To show the effectiveness of our multi-task learning schema (\textbf{Q1}), we compute the correlations of the change in validation metric for each task during training, and compare the performance between: our training schema, training each task individually and multitask training with uniform task sampling. For any two tasks $(A\not = B)$ with metric $\tau_A$ and $\tau_B$, we compute $\rho_{A\to B} \equiv \text{corr}(\delta \tau_A,\delta \tau_B)$ according to the changes in $\tau_A$ and $\tau_B$ after each epoch trained for task $A$. The heatmap for $\rho$ is provided in Figure \ref{fig:corr_and_radar}, where we observe positive correlations among almost all tasks during the training. Specifically, the correlation between the \texttt{substitute} task and all other tasks are high. This indicates that our approach is benefiting from the \textsl{propagation rule}. The radar map in Figure \ref{fig:corr_and_radar} shows that the proposed training schema uniformly outperforms individual task training and the multi-task training under uniform task sampling. 

\subsection{Knowledge Completion and Downstream Tasks Performance}
\label{sec:downstream}

\begin{table}[htb]
\footnotesize
    \centering
    \begin{tabular}{c|ccccc}
    \hline
        Model & \textbf{FM}  & \textbf{BPR} & \textbf{prod2vec} & 
        \textbf{triple2vec} & \textbf{Our approach} \\ \hline 
        \textbf{Hit@10} (a11) & 4.24 & 7.65 & 6.39 & \underline{\textsl{11.30}} & \textbf{\textsl{13.72}} \\
        \textbf{NDCG@10} (a12) & 1.85 & 3.17 & 2.26& \underline{\textsl{4.83}} & \textbf{\textsl{5.79}} \\ \hline
    \end{tabular} 
    
    \vspace{0.3cm}
    
    \begin{tabular}{c|cc|cc}
    \hline
        Task & \multicolumn{2}{c}{\textbf{Encountered queries}} & \multicolumn{2}{c}{\textbf{New queries}} \\ 
        \hline
        Metric & \textbf{R@10} & \textbf{MAP@10} & \textbf{R@10} & \textbf{MAP@10} \\ 
        & (a13) & (a14) & (a15) &  (a16)  \\
        \hline 
        \textbf{TransE} & 8.74  &  3.26 &  5.11 & 2.62  \\
        \textbf{TransH} & 10.43  & 4.28  & 6.85  & 2.79  \\
        \textbf{TransR} & 15.82  & 6.77  & 10.33  &  4.09 \\
        \textbf{TransD} &  13.69 & 6.17  & 9.50  & 4.24  \\
        \textbf{RESCAL} &  7.71 &  2.98 & 4.25  & 1.93  \\
        \textbf{DistMult} & 19.72  & 8.43  & 11.71  & 5.02  \\
        \textbf{ComplEx} & \underline{\textsl{21.58}}  & \underline{\textsl{10.04}}  & \underline{\textsl{12.46}}  & \underline{\textsl{5.15}}  \\
        \textbf{Our approach} & \textbf{\textsl{30.99}}  & \textbf{\textsl{14.46}}  & \textbf{\textsl{22.71}}  &  \textbf{\textsl{9.53}} \\ \hline 
    \end{tabular}
    \caption{Testing performance (in \%) on next-impression recommendation and the search ranking task for queries that have encountered in training and new queries. The results are averaged over three runs.}
    \label{tab:result-search}
\end{table}

%


From Table \ref{tab:result-knowledge-completion}, we see that the proposed approach outperforms all baselines in all tasks for knowledge completion, even when the KG embedding methods are enhanced with the pre-trained product knowledge graph (\textbf{Q4}). The fact that KG embeddings have subpar results when not using PRG suggests they rely heavily on well-established facts which are absent in the raw dataset (\textbf{Q3}). Notice that our approach exceeds in completing \texttt{complement} and \texttt{view} by significant margins. Since \texttt{complement} and \texttt{view} are the two relations which contain the most sophisticated semantics, we believe that our solution of using distributed representation performs better in capturing the richer relational semantics. 

The results in Table \ref{tab:result-search} suggest that the PKG embeddings learned by our approach can also benefit downstream tasks such as search ranking and recommendation (\textbf{Q2}). In the search ranking task, our approach significantly surpasses baselines on both encountered queries and new queries. It is worth pointing out that for the next-impression prediction task, we also outperform the baseline models which are specially designed for recommendation.

\section{Conclusion and Future Work}
\label{sec:conclusion}
We fully characterize the product knowledge graph and systematically compare it with the ordinary knowledge graph. To effectively learn PKG embedding with generic e-commerce dataset, we propose a self-attention-enhanced distributed representation learning method with an efficient multi-task training schema. The empirical results on the real-world data show that our approach outperforms KG embedding baselines in knowledge completion and delivers promising outcomes in downstream search ranking and recommendation. In the future, we will explore incorporating customer and customer knowledge (e.g. demographic information) into PKG to construct the customer-product knowledge graph that can stand out as the backbone for personalized e-commerce services.

\bibliographystyle{ACM-Reference-Format}
\bibliography{WSDM-reference}

\end{document}